\documentclass{article}

\usepackage[preprint]{neurips_2018}

\usepackage[utf8]{inputenc} 
\usepackage[english]{babel}
\usepackage[T1]{fontenc}    
\usepackage{hyperref}       
\usepackage{url}            
\usepackage{booktabs}       
\usepackage{amsfonts}       
\usepackage{nicefrac}       
\usepackage{microtype}      

\usepackage{amsthm}
\usepackage{amsmath,units}
\usepackage{pgfplots}
\usepackage{amsmath} 
\usepackage{tikz}
\usepackage[sc]{mathpazo} 
\usepackage[T1]{fontenc} 
\usepackage{microtype} 
\pgfplotsset{compat=1.6}
\usepackage{mathrsfs}
\usepackage{multicol} 
\usepackage[hang, small,labelfont=bf,up,textfont=it,up]{caption} 
\usepackage{booktabs} 
\usepackage{float} 
\usepackage{hyperref} 

\usepackage{etoolbox}
\patchcmd{\thebibliography}{\section*{\refname}}{}{}{}

\usepackage{lettrine} 
\usepackage{paralist} 

\usepackage{titlesec} 
\usepackage{fancyhdr} 
\usepackage{graphicx}
\restylefloat{figure}
\usepackage{url}
\usepackage{amssymb}
\newcommand{\jeff}[1]{\noindent\underline{\textbf{#1}}}

\theoremstyle{definition}
\newtheorem{definition}{Definition}[section]
\theoremstyle{remark}

\usepackage{algorithm}
\usepackage[noend]{algpseudocode}

\title{Bilingual is At Least Monolingual (BALM):\\ \vspace{2mm} \large A Novel Translation Algorithm that Encodes Monolingual Priors}

\author{%
  Jeffrey Cheng\thanks{Please find my most recent contact information at \texttt{http://jeffreyscheng.com/}.} \\
  Department of Computer Science\\
  University of Pennsylvania\\
  Philadelphia, PA 19104\\
  \texttt{jeffch@seas.upenn.edu} \\
  \And
  Chris Callison-Burch\\
  Department of Computer Science\\
  University of Pennsylvania\\
  Philadelphia, PA 19104\\
  \texttt{ccb@cis.upenn.edu}\\
}

\begin{document}

\maketitle

\begin{abstract}
State-of-the-art machine translation (MT) models do not use knowledge of any single language's structure; this is the equivalent of asking someone to translate from English to German while knowing neither language.  \textbf{BALM} is a framework incorporates monolingual priors into an MT pipeline; by casting input and output languages into embedded space using BERT, we can solve machine translation with much simpler models.  We find that English-to-German translation on the Multi30k dataset can be solved with a simple feedforward network under the \textbf{BALM} framework with near-SOTA BLEU scores.
\end{abstract}

\section{Introduction}

We motivate this research with a problem, an observation, and a technology.
\begin{enumerate}
    \item \jeff{The problem domain}:
    
    Machine translation (MT) requires many pairs of parallel texts (also known as bilingual corpora): pairs of documents that are sentence-wise identical in meaning but written in different languages.  Since translation is a variable-length learning problem in both input and output, it requires a large quantity of data.  Machine translation on pairs of languages where parallel texts are scarce is an open problem (e.g. very few documents are written in both Haitian Creole and Sanskrit, so translation between the two is hard).
    
    \item \jeff{A observation about current solutions}:
    
    It would be unreasonable -- ludicrous, even -- to ask a person who speaks neither English nor German to perform English-German translation.  And yet most MT algorithms incorporate no prior knowledge of either its input or output language at any level (syntax or semantics); instead, these MT algorithms learn the translation problem directly.  This goes against the conventional wisdom of building inductive biases based on domain expertise (e.g. translational invariance, rotational invariance, hierarchical encoding) into models.
    
    \item \jeff{An underutilized technology}:
    
    It is hypothesized that the bidirectional encoder representations from transformers (BERT) algorithm allows practitioners to incorporate prior knowledge about a language by encoding sentences as fixed-length embeddings.  If this hypothesis is true, then BERT can leverage the observation about how humans perform translation in order to address the MT problem posed.  However, no work has proven whether BERT's mean-pooled fixed-length encodings can actually serve as a sentence embedding.
\end{enumerate}

Suppose we want to translate between two languages $A$ and $B$.  Further suppose that there are large quantities of written text in both of these languages, but there are very few parallel texts in languages $A$ and $B$.  Current MT algorithms are unable to solve this problem, despite the fact that the vast majority of pairs of languages fall under this description.

Our premise is to develop a novel algorithm -- the \textbf{Bilingual-is-At-Least-Monolingual model} (\textbf{BALM}) -- which uses BERT to incorporate prior knowledge about language $A$ and language $B$ independently.  This encoding of prior knowledge allows \textbf{BALM} to learn the translation problem as a fixed-length mapping problem (which is easier and more data-efficient).  We now motivate our concept more deeply.

\subsection{The Problem Domain: MT Restricts Choice of Model}

MT is difficult in part because natural language is variable-length: the number of tokens in sentences varies greatly.  Most classification models (logistic regression, decision trees, support vector machines, feedforward neural networks) have a rigid API that only allows for fixed-length inputs and fixed-length outputs.  Thus, they are not only unable to learn MT; they are unable to compute even a single-forward pass over the data.

Certain architectures are designed for variable-length problems, notably recurrent neural nets (RNNs) and recursive neural nets.  However, even RNNs cannot learn MT directly since their API is still not flexible enough.
\begin{itemize}
    \item RNNs can map from variable-length inputs to fixed-length outputs.
    \item RNNs can map from fixed-length inputs to variable-length outputs.
    \item RNNs can map from variable-length inputs to variable-length outputs \textbf{iff} the input and output have the same length.
\end{itemize}
There are no atomic models that directly allow for arbitrarily variable-length inputs and outputs.  Since MT has natural langauge as both input and output, there are therefore no atomic models that can directly solve MT.

MT practitioners are thus forced to use rigid compound models that either use recurrence or enforce an awkward sequence length maximum.  We would like to relax the conditions of the MT problem to allow simpler models, such as feedforward neural networks.

\subsection{Related Work in NLP and NMT}

Neural machine translation (NMT) was popularized by encoder-decoder networks' impressive performance between English and French in 2015.\cite{bengio-ed}

Bengio et al's seq2seq architecture got around the variable-length problem by a simple composition of two attentioned RNNs: an encoder to embed sentences into a fixed-length \textbf{thought vector} and a decoder to interpret the thought vector as language.  The attention mechanism allowed the models to recognize nonconsecutive patterns in sequence data directly without sole reliance on memory gates.  Seq2seq represented a clean improvement from prior efforts in statistical machine translation because of its compactness and because of its end-to-end differentiability.

Shortly after Bengio et al's success with seq2seq, NMT literature noted two fundamental challenges with using recurrent models for translation. 
\begin{enumerate}
    \item Hochreiter et al found that each variable-length touchpoint within recurrent models exacerbates their susceptibility to vanishing gradients since the path along the computation graph from the end loss to early weights increases linearly in the sequence length.  The chain rule is multiplicative along the gradient path; therefore, the gradient may decay exponentially, which prevents efficient training.\cite{vanishing-gradient}  Since MT has two such variable-length touchpoints (variable-length inputs and variable-length outputs), recurrent models for MT are difficult to tune and train reliably.
    \item Neural networks require a large quantity of data -- even simple problems like MNIST require 60K examples and several epochs in order to converge.  MT is particularly a noisy domain.  Koehn and Knowles note that the combination of these two factors makes NMT is an extremely data-hungry problem setup.  \cite{data-hungry}
    
    Koehn and Knowles' observation is exacerbated by the fact that most pairs of languages are not like the English-French translation problem solved by Bengio et al.  English and French are both extremely widely-used languages, and the pair has an enormous number of parallel texts.  Most pairs of languages have few parallel texts, and the data requirement imposed by such a model for universal translation scales quadratically in the number of languages.

    For example, anyone can find a public corpus of Romanian text on the order of billions of words with a quick search.\cite{Romanian-text}  However, the largest parallel text source between Romanian and another language is only about 1 million tokens. \cite{Parallel-Corpora}
\end{enumerate}

Further developments in attention led to Google Brain's development of the transformer, which currently holds state-of-the-art results in machine translation between English-German and English-French.\cite{transformer}

Transformers utilize an encoder-decoder scheme using three kinds of multi-headed attention (encoder self-attention, decoder self-attention, and encoder-decoder attention).  Transformers are a fixed-length attention-based model, which reduces the vanishing gradient problem.  However,  even the smallest pre-trained English transformer models have over 110 million parameters -- the size of this model \textbf{worsens} the data-hungriness problem.

Several works have attempted to design algorithms that artificially augment the sizes of NLP and MT datasets.
\begin{itemize}
    \item Sennrich et al developed backtranslation, a semi-supervised algorithm that concatenates monolingual utterances to bilingual corpora by using NMT to impute a translation.  Backtranslation demonstrates modest increases in BLEU.  However, neural nets under the backtranslation setup still do not learn the structure of any single language; the imputed examples are still used for directly learning the translation task.\cite{sennrich}
    \item Burlot et al found that variants of backtranslation and GAN-based data augmentation can increase the size of the dataset slightly without compromising the generalizability of the neural net.  However, since learning curves are extremely sublinear with respect to duration, these data augmentation schemes provide small, diminishing returns.\cite{burlot}
    \item Sriram, Jun, and Satheesh developed cold fusion (an NLP technique, contrasted with deep fusion), which performs simultaneous inference and language modeling in order to reduce dependence on dataset size.  This is the exact inductive bias that humans use and is the jumping point for our work.  However, cold fusion is unable to extend from monolingual inference to MT.  This is likely because at the time of Sriram et al's writing (August 2017), no general sentence embedding algorithm existed.  As explained later, we now bypass this constraint with Google's BERT algorithm.\cite{cold-fusion}
\end{itemize}
None of these algorithms adequeately addresses our first observation that MT given natural language understanding in a single language is a much easier learning task than directly learning MT.

Therefore, since data quantity is a performance bottleneck and and since bilingual training data for translation models is typically difficult to obtain, we would like to pre-train the models as much as possible by learning efficient representations using \textbf{monolingual data} before attempting translation.  Kiros et al make progress towards pretrained monolingual language modeling for MT by learning multimodal embeddings on words.\cite{multimodal}

We will progress in the same vein but jump directly to pre-training sentence embeddings rather than word embeddings by using the encoding prowess of transformers.  This approach could have the added advantages of context-based disamguation and better understanding of syntactical arrangement.

\subsection{An Underutilized Technology: BERT for Translation}

Deep bidirectional transformers for
language understanding (BERT) is an extremely popular pre-trained transformer model that is supposedly able to encode sentences as fixed length vectors using only a monolingual corpus; this represents an improvement over previous language embedding technologies such as ELMO, which could only embed words into vectors.\cite{bert}

BERT uses the encoder stack of a transformer architecture to return an context-driven embedding for each word in a sequence -- practitioners have found that taking the mean pool of the word embeddings in a sentence functions well as a sentence embedding.  \cite{baas}  However, BERT's sentence embedding property has never been verified with an autoencoder.

We note that if BERT is able to embed sentences into fixed-length vectors, the problem of translation no longer has the difficulty of being a variable-length input problem.  Similarly, if we can find an inversion of the BERT model, the MT problem will no longer be a variable-length output problem.  Combining these two hypothetical successes would cast MT into a fixed-length learning problem, which can be solved with any arbitrary classification algorithm.  We would then be able to use much simpler models than attentioned RNNs (such as feedforward neural nets or even logistic regression).

\subsection{Research Goals}

\begin{enumerate}
    \item Create an English sentence autoencoder using a pre-trained English BERT as the encoder and a newly initialized recurrent decoder.
    \begin{itemize}
        \item If the autoencoder can reconstruct sentences with high fidelity, we will have verified that mean-pooling over BERT's word embedding does create sentence embeddings.  We will also have found a useful English thought-space that can be used for transfer learning.
        \item It is still interesting if the autoencoder fails since this disproves the running assumption of many practitioners that a mean-pooled BERT output acts as a sentence embedding.  We would then attribute BERT's positive performance as a transfer learning tool to its ability to learn domain-specific embedding features but an inability to compress all sentence features into a single vector.
    \end{itemize}
    \item Conditional on a successful autoencoder, create a German-English translation model using German BERT as an encoder, a translation model mapping between English and German thought-spaces, and the aforementioned BERT-inverting decoder.
    \item Conditional on a successful translator, compare its learning curve with SOTA models and check for convergence with shorter duration (fewer minibatches / epochs).  If \textbf{BALM} is successful here, then it shows promise as an algorithm for MT in pairs of languages with few bilingual corpora. 
    \item Conditional on a successful translator, check for reasonably good translations as evaluated by bilingual evaluation understudy (BLEU).\cite{bleu}  Similarly, conditional on a successful translator, check for qualitatively good translations on in-sample and out-of-sample sentence examples.  If \textbf{BALM} is successful here, then it shows promise as an algorithm for general MT as a substitute for current SOTA methods.
\end{enumerate}

\section{Model Descriptions}

We begin with a non-standard definition that will clarify the model premises.

\theoremstyle{definition}
\begin{definition}{(Thought-space)}
For a language $L$, a thought-space $S_{L, k}\subseteq \mathbb{R}^k$ is a $k$-dimensional embedding of sentences in language $L$ for some fixed $k\in \mathbb{N}$.
\end{definition}

The key insight of seq2seq learning is that by learning an intermediate thought-space, the overall sequence-to-sequence task becomes easier since the subproblems are learnable by RNNs.  Our insight here is that we can make machine translation significantly easier by learning 2 intermediate thought-spaces.

Suppose we want to translate German $\rightarrow$ English, with $L_{\text{English}}$ and $L_{\text{German}}$ representing the formal languages.  We learn the two intermediate thought-spaces $S_{\text{English}, k}$ and $S_{\text{German}, k}$ -- each language has its own embedded space.  We thus construct the \textbf{Bilingual-is-At-Least-Monolingual} (\textbf{BALM}) model for German-to-English translation with the following three submodules:
\begin{enumerate}
    \item A German BERT encoder learns to embed German natural language into a fixed-length embedding.  Its outputs are German ``thoughts.''  \textbf{This is learnable by monolingual datasets}.  In fact, one can easily download a pretrained BERT model that does this; no bilingual corpora is necessary.  Formally,
    \begin{align*}
        B_{\text{German}}: L_{\text{German}}\rightarrow S_{\text{German}, k} \tag{Encodes from German for fixed $k$}
    \end{align*}
    \item A feedforward neural net translates from fixed-length German thoughts into fixed-length English thoughts.  Note that since this is a fixed-length problem, the learning task is significantly easier in this step and should require less data.
    \begin{align*}
        F_{\text{German}\rightarrow\text{English}}: S_{\text{German}, k}\rightarrow  S_{\text{English}, k} \tag{Translates German thoughts into English thoughts}
    \end{align*}
    \item A recurrent English decoder learns to reconstruct English natural language from fixed-length English thoughts.  \textbf{This is learnable by monolingual datasets}.
        \begin{align*}
        B_{\text{English}}^{-1}: S_{\text{English}, k}\rightarrow L_{\text{English}} \tag{Decodes into English for fixed $k$}
    \end{align*}
\end{enumerate}
Under this framework, translation has been stripped of its variable-length property -- we only need to train one atomic model with the bilingual parallel texts!

We must first pre-train $B_{\text{German}}$ and $B^{-1}_{\text{English}}$.  We will take the former for granted since there are published pre-trained German BERT models.  We will train $B^{-1}_{\text{English}}$ with an autoencoder, then transfer-learn by using the pre-trained weights of $B_{\text{German}}$ and $B^{-1}_{\text{English}}$ in a full translation model.

\subsection{Autoencoder Model}

 We utilize a pre-trained BERT embedding model $B_{\text{English}}$.
\begin{align*}
    B_{\text{German}}:& L_{\text{English}}\rightarrow S_{\text{English}, k} \tag{Encodes from English for fixed $k$}
    \intertext{In order to build an autoencoder with a BERT encoder, we need to be able to \textit{invert} the function $B_{\text{German}}$.}
    B_{\text{English}}^{-1}:& S_{\text{English}, k} \rightarrow L_{\text{English}}\tag{Decodes into English for fixed $k$}
\end{align*}
We implement $B_{\text{English}}^{-1}$ as an RNN with gated recurrent units (GRUs).  We choose not to use vanilla RNNs due to concerns about vanishing gradients -- natural language frequently reaches sequences of lengths greater than 50.  We choose the GRU over the LSTM because its API is more compatible with the BERT encoder; BERT returns one hidden state; GRUs recur with one hidden state while LSTMs recur with two.  

The autoencoder $A$ is thus $\boxed{A=B_{\text{English}}\circ B_{\text{English}}^{-1}}$.  Note that $A(x)\approx x$.

\subsection{Translation Model}

Once the autoencoder has converged, we copy $B_{\text{English}}^{-1}$ as a pre-trained decoder and use a pre-trained German BERT model $B_{\text{German}}$.  This allows us to transfer-learn from the language modeling task to the translation task.  The full \textbf{BALM} translation model is:
\begin{align*}
    B_{\text{German}}:& L_{\text{German}}\rightarrow S_{\text{German}, k}\\
    F_{\text{German}\rightarrow\text{English}}:& S_{\text{German}, k}\rightarrow  S_{\text{English}, k}\\
    B_{\text{English}}^{-1}:& S_{\text{English}, k}\rightarrow L_{\text{English}}
\end{align*}

The translator $T$ is thus the composition $\boxed{T=B_{\text{German}}\circ F \circ B_{\text{English}}^{-1}}$.

Note that while the translator is training, the feedforward network $F_{\text{German}\rightarrow\text{English}}$ is the only model being trained from initialization.  We merely fine-tune the two transferred models $B_{\text{German}}$ and $B_{\text{English}}^{-1}$.  

Given the two compound models that need to be trained (autoencoder $A$ and \textbf{BALM} translator $T$), we proceed to our MT methodology.

\section{Methodology}

\subsection{Data source}

We utilize Elliott et al's Multi30k translation dataset\footnote{\url{https://github.com/multi30k/dataset}}, a multilingual extension of the Flickr30k image-captioning dataset.\cite{multi30k}  Each image has an English-language caption and a human-generated German-language caption.  We depict one such example below.

\vspace*{0.5cm}
\begin{figure}[h!]
\includegraphics[width=\textwidth]{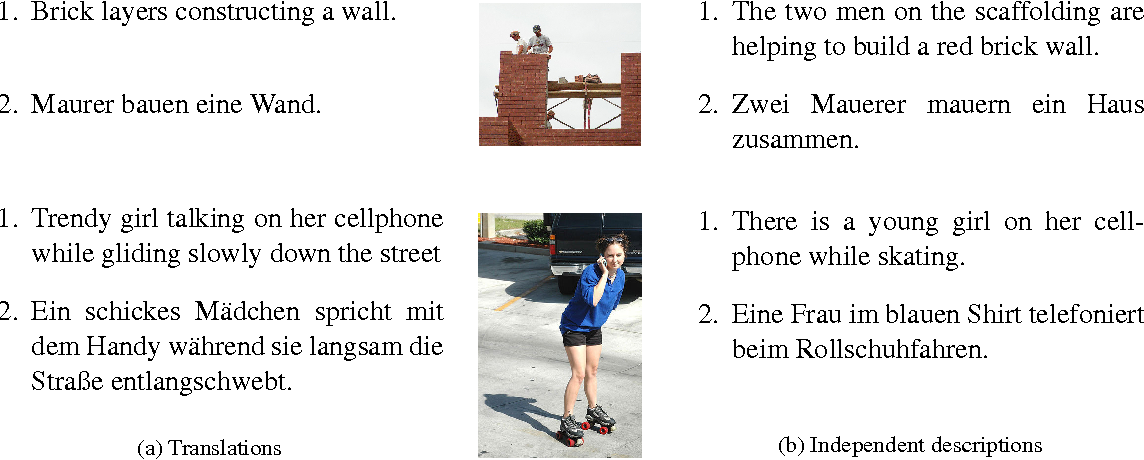}
\caption{We see two images here.  (Right) Each image has an independently written caption in both English and German.  (Left) Each independent caption has a corresponding translation in the dataset.  Image courtesy of Elliott et al.}
\end{figure}
\vspace*{0.5cm}
By the nature of image captioning, this dataset is biased towards static structures (e.g. "This is a white house.") and animals doing things ("The brown dog is drinking from the bowl.").  Notably, there are no questions, commands, or abstract statements in the data; thus, we should expect that our models are only able to create good representations of descriptive text.

We load this dataset using the builtin function within torchtext, an NLP extension of Facebook's PyTorch framework. Natural language cannot be directly fed into an encoding model, so we must:
\begin{enumerate}
    \item Tokenize the text.
    \item Convert the text to categorical id values.
    \item Initialize one-hot vectors.
    \item Embed the vectors into a low-dimensional space for learning.
\end{enumerate}

We utilize a hybrid ecosystem of standard \textit{PyTorch}\footnote{\url{https://pytorch.org/docs/stable/index.html}}, the NLP extension \textit{torchtext}\footnote{\url{https://torchtext.readthedocs.io/en/latest/}}, and \textit{pytorch-pretrained-bert}\footnote{\url{https://github.com/huggingface/pytorch-pretrained-BERT\#usage}} a library within the PyTorch ecosystem for pre-trained transformers developed by huggingface.ai.  The diagram depicts the pipeline infrastructure and the dimensionality of the flow.

\begin{figure}[H]
\begin{center}
\includegraphics[width=\textwidth]{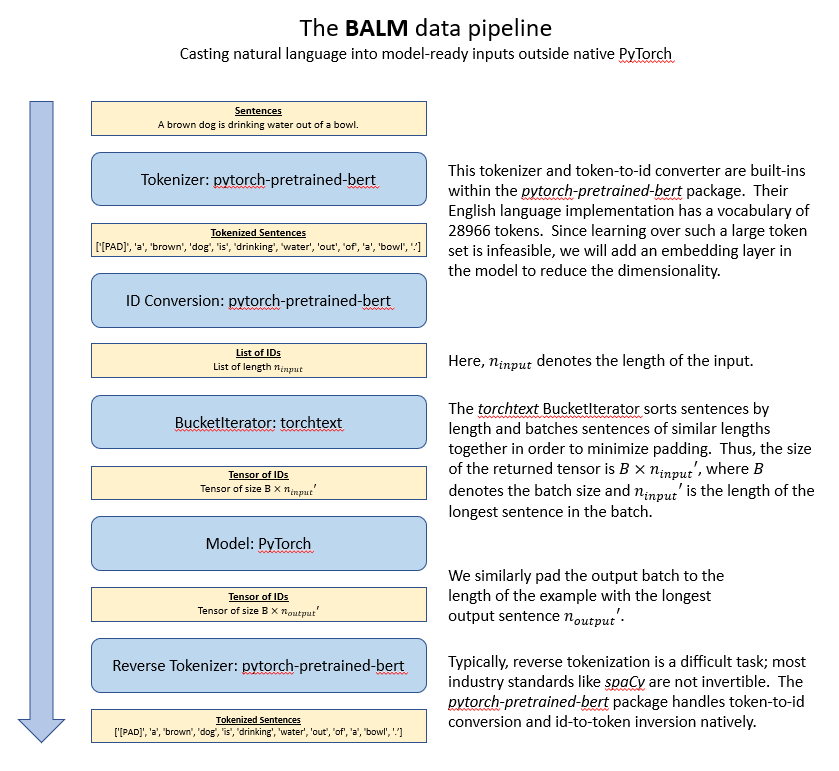}
\caption{Three libraries transform natural language data into model-ready batch-wise tensors. Original flow chart diagram and original descriptions.}
\end{center}
\end{figure}

We now expand out the model and explain the forward passes of the autoencoder and the translator.

\subsection{Model implementation}

\subsubsection*{Autoencoder Implementation}

We implement the \textbf{BALM} autoencoder model in native PyTorch with the following modules:
\begin{itemize}
    \item The pre-trained English BERT encoder.  This is downloaded from the \textit{pytorch-pretrained-bert} package and has 110 million parameters.  We allow the gradient updates of the autoencoder to backpropagate through the pre-trained BERT model in order to fine-tune its embedding.  Note that we are not using the encoder in the \textbf{BALM} translation model; the fine-tuning is solely to improve learning outcomes for the GRU decoder model.
    \item A mean pool layer, implemented with \textit{torch.mean}.  The BERT encoder has 12 self-attention layers, each of which outputs a hidden state.  We could theoretically learn over all 12 hidden states since this would capture the entire set of low-level and high-level features embedded by the transformer.  However, learning over such a rich feature space would present memory issues for the GPU; thus, we take only the last one to obtain an vector of size 768 for each token.  We then mean-pool across the dimension of sentence length to get a sentence embedding of size 768; this operation is optimized by PyTorch's CUDA interface.
    \item The English GRU decoder layer.  This is implemented as a single-layer RNN with gated recurrent units and hidden dimension equal to that of BERT's encoding.  In order to produce an output at a given timestep $t$, the GRU takes the hidden state at $t$ and passes it through a linear layer of output size equal to the vocabulary length (28996).
    \item A word embedding layer (not depicted below).  We implement teacher-forcing on the GRU decoder in order to learn tail-end patterns early in training.  In order to teacher-force correctly, the incoming target words have the same dimensionality as the GRU's hidden layers.  Thus, we learn a custom word embedding, implemented with \textit{nn.Embedding}.
\end{itemize}

The autoencoder's forward pass thus uses 3 trainable modules: the word embedding linear layer, the transformer encoder stack from BERT, and the GRU decoder. The embedding and the decoder are trained from initialization.  The sequence of layers and rationale for architecture choices are depicted in the following flow diagram.  

\begin{figure}[H]
\begin{center}
\includegraphics[width=\textwidth]{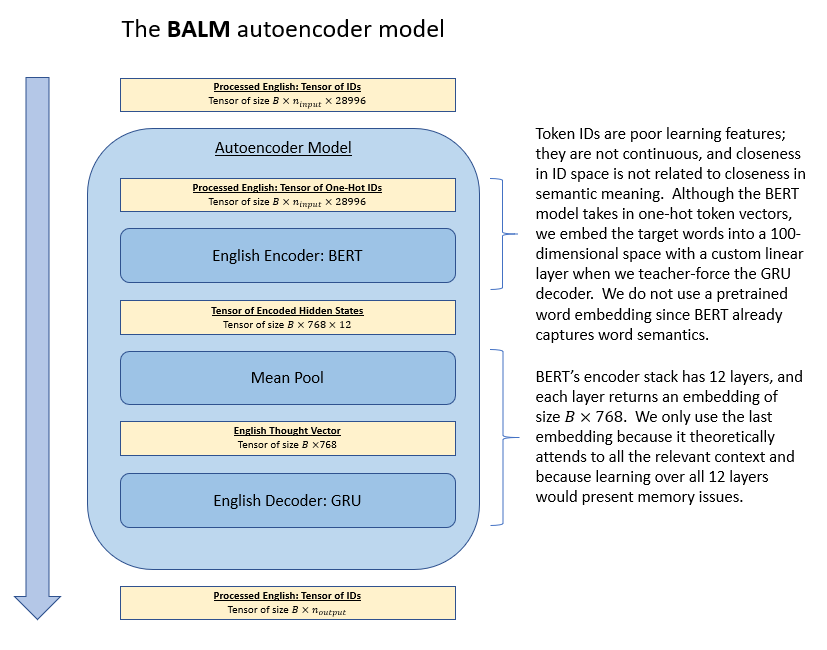}
\caption{Original flow chart diagram and original descriptions.}
\end{center}
\end{figure}

\subsubsection*{Translator Implementation}

We implement the translator with the same framework as the autoencoder, substituting the English BERT encoder for a German BERT encoder.  We also add in an intermediate module to learn the desired function $F_{\text{German}\rightarrow\text{English}}$ mapping between the thought-spaces of the two languages.
\begin{itemize}
    \item The pre-trained German BERT encoder.  This is downloaded from the \textit{pytorch-pretrained-bert} package and has 110 million parameters; the specific model used in this experiment is actually a multilingual BERT encoder that can taken in tokens in English, French, or German.  We fine-tune the model by allowing gradient updates.
    \item A mean pool layer, implemented with \textit{torch.mean}.  The rationale for this module is the same as the rationale for the mean pool layer in the autoencoder.
    \item The feedforward network representing a thought translator (implemented with two \textit{nn.Linear} units and ReLU activations).  This neural net's purpose is to learn the fixed-length function $F_{\text{German}\rightarrow\text{English}}$.  It has architecture $768\times 768\times 768$; its input and output dimensions are fixed by the output dimension of the BERT encoder.
    \item The English GRU decoder layer.  This is transfer-learned from the autoencoder.  We allow the gradient updates of the autoencoder to backpropagate in order to fine-tune the model for translation.
    \item A word embedding layer (not depicted below) for teacher-forcing.
\end{itemize}

The translator's forward pass thus uses 4 trainable modules: the word embedding linear layer, the transformer encoder stack from BERT, the feedforward network, and the GRU decoder. The sequence of layers and rationale for architecture choices are depicted in the following flow diagram. 

\begin{figure}[H]
\begin{center}
\includegraphics[width=\textwidth]{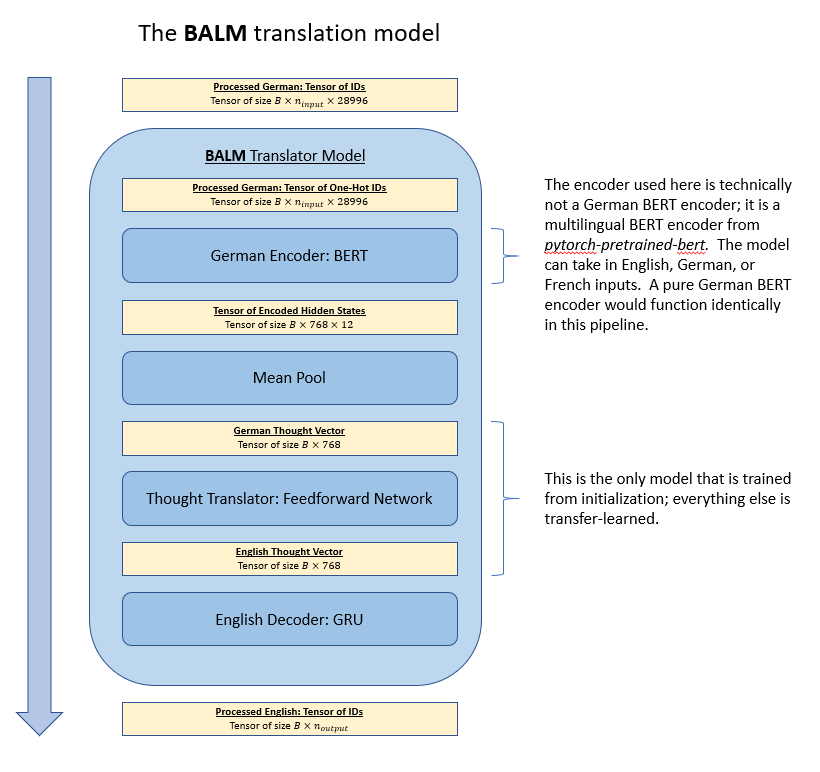}
\caption{Original flow chart diagram and original descriptions.}
\end{center}
\end{figure}

\newpage
We tuned the hyperparameters of the two models to the following constraints:
\begin{itemize}
    \item Minimize loss on the validation set.
    \item Minimize iteration-to-iteration variability.
    \item Keep total memory allocation under 10GB in order to run the model on a standard Nvidia GeForce GTX 1080 GPU.
\end{itemize}

The final tuned hyperparameters are listed below.  Surprisingly, the models seemed to be relatively insensitive to choices of hyperparameters.  We attribute this to the large amount of pretraining in the submodules for both the autoencoding and the translating tasks; given a good ``seed'' embedding from a careful selection of hyperparameter during the pre-training phase, perhaps the actual task itself because relatively invariant to hyperparameter tuning.

\begin{figure}[H]
\begin{center}
\includegraphics[width=\textwidth]{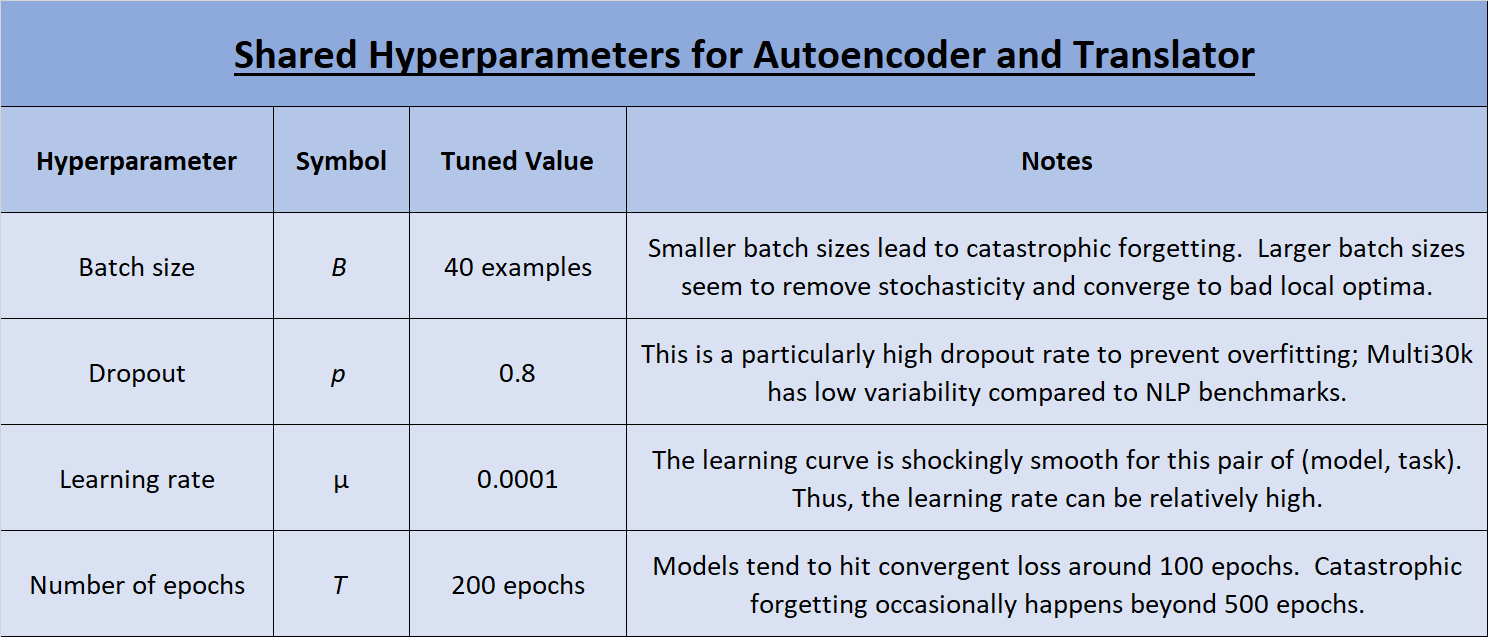}
\caption{The final tuned hyperparameters of both the autoencoder setup and the translator setup.  They are the same for both pipelines for consistency.}
\end{center}
\end{figure}

Given the above hyperparameters, we run the autoencoder model and the translator model in sequence.  Each model's training run of 200 epochs took roughly 20 hours to run: a relatively quick convergence for an NLP task.

\section{Results}

\subsection{Autoencoder Model}

\subsubsection{Learning Curve}

\begin{figure}[H]
\begin{center}
\includegraphics[width=0.5\textwidth]{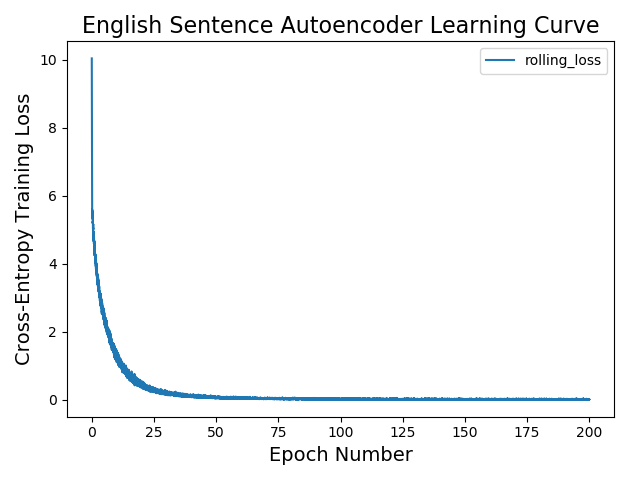}
\end{center}
\end{figure}

The learning curve of the autoencoder has several striking features.
\begin{itemize}
    \item The convergence is extremely fast.  Many NLP tasks take several hundred epochs to converge; the elbow of this learning curve is around 15 epochs.
    \item The convergence is extremely smooth.  Given how noisy natural language is as a domain, this is surprising.
    \item The convergence of the learning curve is right at the Bayes error.  Random guessing over 28996 classes has an expected cross entropy loss of:
    \begin{align*}
        E[L(y, \hat{y})]&\geq L(y, E[\hat{y}])\tag{By Jensen's Inequality; $L$ is concave in $\hat{y}$}\\
        &=-\sum_{z=0}^{28995} \mathbb{I}(z=y)\ln E[\hat{y}_i]\tag{By definition of cross-entropy}\\
        &=-\ln\frac{1}{28996}\approx \boxed{10.27} \tag{Exactly one indicator is one}
    \end{align*}
    We find that the cross-entropy loss converges to an average of 0.012 after epoch 180.  This is the equivalent of the model always answering with the correct token with an assigned probability of $e^{-0.012}\approx 0.988$.  
\end{itemize}

\subsubsection{BLEU score}

While a low cross-entropy loss is a signal of a good model and is conveniently differentiable, it doesn't directly give us the quality of reconstructions.  We turn to the bilingual evaluation understudy (BLEU) score for a more holistic measure of the autoencoder's reconstruction.  The BLEU score measures the ``adequacy, fidelity, and fluency'' of proposed translations by measuring the proportion of $n$-grams shared between the proposed translation and the ground-truth translation.\cite{bleu}

The \textbf{BALM} autoencoder achieves a remarkably high BLEU score of $\boxed{0.605}$.

\subsubsection{Qualitative Analysis of Reconstructions}

\noindent Finally, in order to qualitatively interpret the output of a model, we convert the ouput IDs back into tokens.  We utilize the reverse-tokenization builtins from \textit{pytorch-pretrained-bert}.  We observe the autoencoder's outputs for training examples, test examples, handwritten-caption-like examples, and non-caption-like examples.

\begin{figure}[H]
\begin{center}
\includegraphics[width=\textwidth]{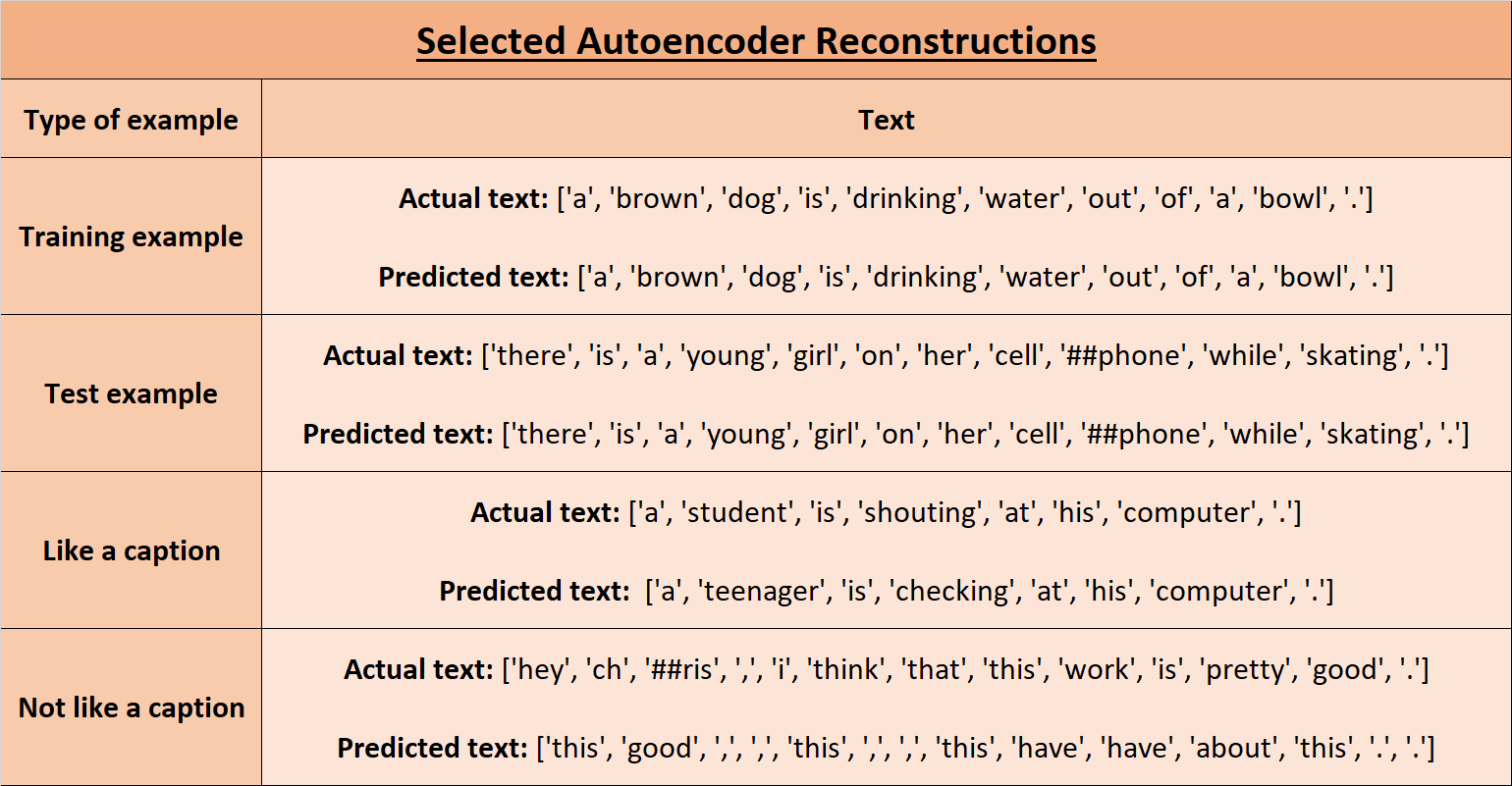}
\caption{Selected examples of the \textbf{BALM} autoencoder's reconstructions.}
\end{center}
\end{figure}

We note that on examples from the Multi30K dataset (train and test), the autoencoder has perfect reconstructions.  For manually-written examples that are somewhat similar to captions, the autoencoder gets the rough syntax and semantically similar words.  On examples that are not like captions (not declarative sentences), the autoencoder is only able to capture the sentiment of the sentence.

Overall, this is exactly what we'd expect of a properly trained autoencoder on the Multi30K dataset.  We take the favorable learning curve, the high BLEU score, and the selected reconstructions as strong evidence that BERT creates good sentence embeddings and that the \textbf{BALM} autoencoder is able to capture the English thought-space.

\subsection{Translation Model}

\subsubsection{Learning Curve}

\begin{figure}[H]
\begin{center}
\includegraphics[width=0.8\textwidth]{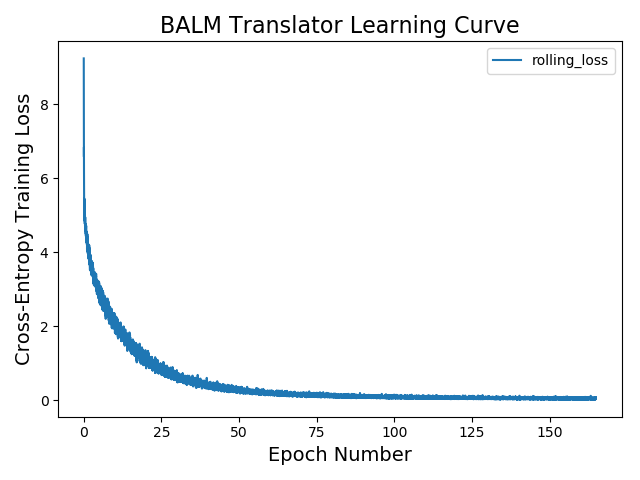}
\end{center}
\end{figure}

We notice similar features in the translator learning curves as compared to the autoencoder's.
\begin{itemize}
    \item The convergence is extremely fast.  The translator converges just a hair slower than the autoencoder.
    \item The convergence is extremely smooth.  Again, the translator's loss is slightly noisier than the autoencoder.
    \item The convergence of the learning curve is right at the Bayes error.  Recall that random guessing over 28996 classes has an expected cross entropy loss of at least 10.27.  After epoch 140, the translator model has an average loss of 0.014.
\end{itemize}

\subsubsection{BLEU score}

We again use the BLEU score for evaluation -- this time, we use it as it was intended: for bilingual translation.  We find that the score is $\boxed{0.248}$.  Although this is not nearly as high as our autoencoder score and falls short of the state-of-the-art performance on Multi30K of 0.35, we take this BLEU score as a weak success.  In general, MT practitioners consider a BLEU score above 15\% to indicate that nontrivial learning is happening -- and this is being achieved with a feedforward network!

\subsubsection{Qualitative Reconstructions}

\begin{figure}[H]
\begin{center}
\includegraphics[width=\textwidth]{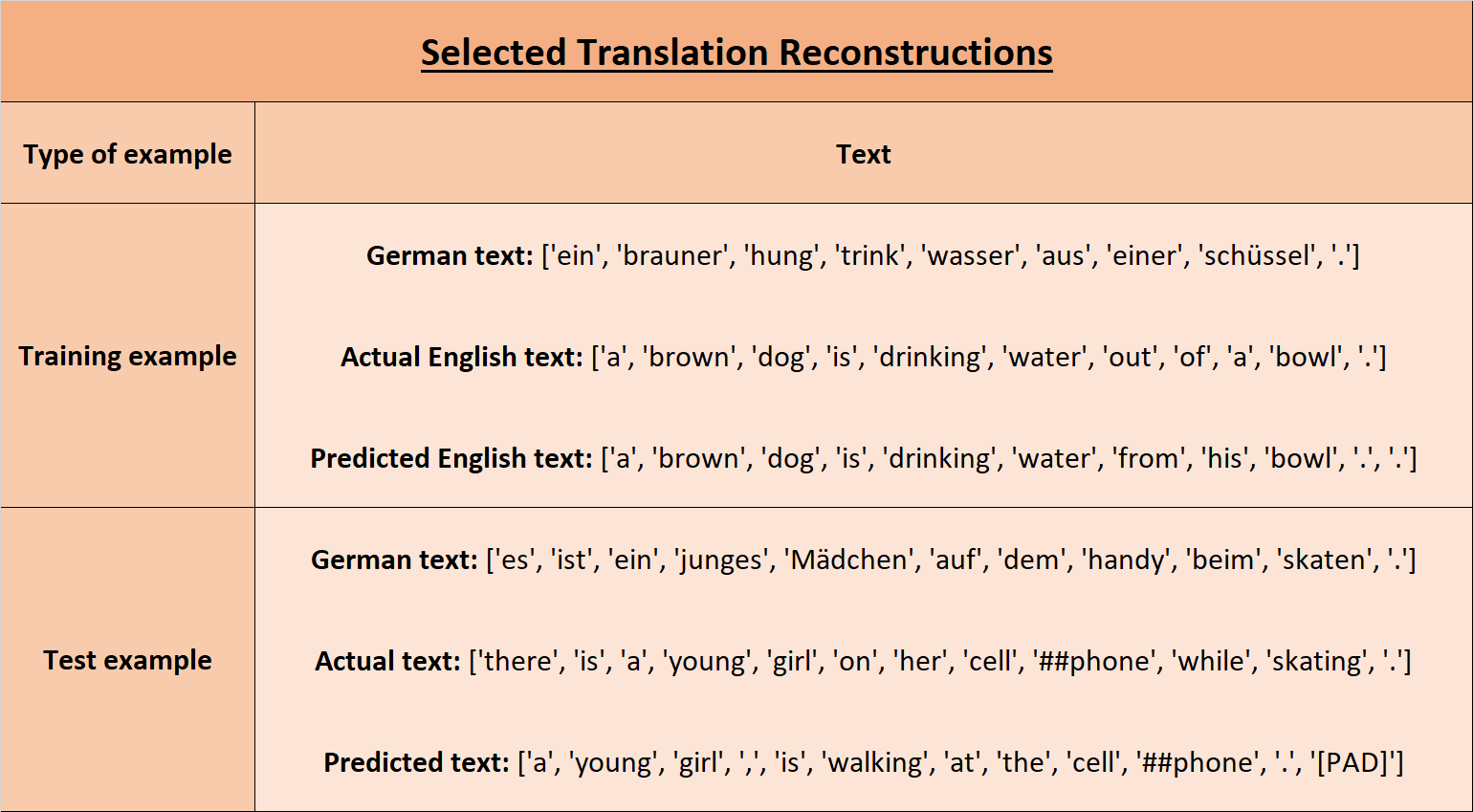}
\caption{Selected examples of the \textbf{BALM} translator's reconstructions.}
\end{center}
\end{figure}

We see that the translation model does relatively well on the training examples.  In fact, it produces the strongest signal of natural language understanding: a correct translation that is synonymous but not identical to the given translation.  However, the model struggles a bit on the test set.  There is clear $n$-gram similarity, but the model misses the key verb and returns a poorly formed sentence.

\section{Conclusion}

We first conclude that NLP practitioners are correct to assume that a simple mean-pool over BERT's word embeddings serve as a rich sentence embedding.  We take the extremely high performance of the BERT-driven sentence autoencoder -- nearly zero cross-entropy loss, an extremely high BLEU score of 0.605, and impressive performance on out-of-dataset examples -- as strong evidence that the English thought-space learned by BERT captures all salient features of the English natural language (at least within the subdomain of image captioning).

This work definitely contradicts Ray Mooney's famous exclamation at the Association of Computational Linguistics (ACL) 2014: ``You can't cram the meaning of a whole \%\&!\$\# sentence into a single \$\&!\#* vector!''

Next, we conclude that BERT embeddings allow extremely complex sequence-to-sequence NLP problems like MT to be solved by extremely simple models like feedforward networks.  A simple transfer learning from an autoencoder and a shallow feedforward network trained on the German-to-English translation task achieves similar training curves, a sub-SOTA but respectable BLEU score of 0.248, and reasonable in-sample reconstructions.

Finally, we conclude that the \textbf{BALM} algorithm does seem to converge faster than both seq2seq and transformer-based MT systems, although its final performance is not as strong.  This bodes well for the initial premise of this research, which was for translating between pairs of languages that have very few parallel bilingual texts.

\section{Impact}

Consider the Haiti earthquake.  Many organizations, such as the United Nations, UNICEF, and the Red Cross, immediately reached out to offer humanitarian aid; however, there are few translators between, say, English and Haitian creole.  Before this crisis, automated translation between frequently spoken languages and Haitian creole did not exist because of a lack of parallel texts.

Microsoft's best efforts reflect positively on the company but poorly on the state-of-the-art of NLP at the time.  Their fast-tracked model involved crowdsourcing parallel texts from universities (e.g. Carnegie Mellon University), websites (e.g. haitisurf.com), and the online Haitian community.  It took Microsoft Research's NLP group 5 days to support Haitian creole on Bing's translator service.\footnote{https://www.microsoft.com/en-us/research/blog/translator-fast-tracks-haitian-creole/}

There are roughly 4500 languages with more than 1000 speakers.\footnote{https://www.infoplease.com/askeds/how-many-spoken-languages}  Now consider a model that could translate between any of the $\binom{4500}{2}=10122750$ pairs of languages with minimal assistance from bilingual corpora.  Translation would no longer depend on a data bottleneck (especially one that's difficult to solve with crowdsourcing given the rarity of speakers in some languages).  Key stakeholders would include humanitarian organizations, any internationally-deployed branch of the military, scholars of dead languages, and language-sensitive content creators.

\textbf{The value of this research is that by better utilizing more plentiful data resources -- monolingual rather than bilingual texts -- we open up the possibility of high-quality machine translation beyond the few pairs of frequently used languages.}

\section{Future Work}

This work has opened up many doors to follow-up analyses and use cases.

\begin{enumerate}
    \item \textbf{Interpretability}
    
    A major issue with neural methods in general is that they are considered opaque: as a semiparametric family of models, it is difficult to analyze what is happening in parameter space.\cite{data-hungry} However, a MT algorithm under the \textbf{BALM} framework has two usable thought-spaces that can be used to visualize any mistakes.  Suppose that a mistake occurs in translation due to the German encoder creating a faulty embedding; we can check this by using a German thought-decoder.  Suppose that a mistake occurs in translation due to the English decoder reconstructing poorly; we can check this by using an English encoder on the poor reconstructor to see if we reproduce the output of the thought translator.  And suppose that a mistake occurs in the thought translator; we can use the English decoder to pinpoint the source of the error.  It would be interesting to see whether certain submodules are more susceptible to certain kinds of errors and whether it's possible to create algorithmic tools for identifying or preventing such mistakes.
    
    \item \textbf{Regularization and Learnability}
    
    We see that a shallow feedforward neural network can function as a reasonable thought translator.  Would a more complex fixed-length model be able to increase the performance of \textbf{BLAM} to SOTA?  A deeper neural network would not be more expressive but could improve the computational learning properties of the system.
    
    Or perhaps the feedforward neural network is too complex and overfits on the data: we saw in the translator that it faithfully reproduced the training examples but struggled with test examples.  Perhaps an extremely simple model like logistic regression would serve as a regularization scheme.
    
    Other forms of regularization could include changing the optimizer (we used Adam by default); some works seem to demonstrate that SGD acts as better regularizer because of its increased stochasticity.  Perhaps combining SGD with \textbf{BALM} would reduce the amount of overfitting.
    
    \item \textbf{Transfer Learning on Different MT Tasks}
    
    We test here only one one dataset: Multi30K.  We do not know if this algorithm will even generalize to other German-to-English translation tasks outside of image captioning.  Furthermore, it would be interesting to see if the target language decoder for one task can be lifted onto a different task altogether; in theory, they should function the same if given the same BERT embedding.
    
\end{enumerate}

\section{Appendix}

\subsection{Links to Research Materials}

Github repository: \url{https://github.com/jeffreyscheng/senior-thesis-translation}

Pre-trained \textbf{BALM} models: \url{https://drive.google.com/drive/folders/1bt8hU24U_Uwn9j7gque2IjGzv4dphz3M?usp=sharing}

\bibliographystyle{plain}
\bibliography{refs.bib}

\end{document}